\pdfoutput=1

\documentclass[11pt]{article}

\usepackage[final]{acl}

\usepackage{times}
\usepackage{latexsym}

\usepackage[T1]{fontenc}

\usepackage[utf8]{inputenc}

\usepackage{microtype}

\usepackage{inconsolata}

\usepackage{graphicx}
\graphicspath{{./figs/}}

\usepackage{amsmath}
\usepackage{amssymb}
\usepackage{hyperref}
\usepackage{geometry}
\usepackage{balance}
\usepackage{enumitem}
\usepackage{breqn}
\usepackage{tikz}
\usepackage{forest}
\usetikzlibrary{trees,positioning,shapes,shadows,arrows.meta}
\title{Biological Sequence with Language Model Prompting: A Survey}

\author{
\bf Jiyue Jiang$^{\heartsuit}$$^{\ast}$,
Zikang Wang$^{\spadesuit}$$^{\heartsuit}$$^{\ast}$,
Yuheng Shan$^{\clubsuit}$$^{\heartsuit}$\thanks{Equal Contribution},
Heyan Chai$^{\heartsuit}$,\\
\bf Jiayi Li$^{\heartsuit}$, 
Zixian Ma$^{\heartsuit}$,
Xinrui Zhang$^{\heartsuit}$,
Yu Li$^{\heartsuit}$\\
$^{\heartsuit}$ The Chinese University of Hong Kong,
$^{\spadesuit}$ The Hong Kong Polytechnic University, \\
$^{\clubsuit}$ National University of Singapore \\
{\tt
\{jiangjy, 1155191449\}@link.cuhk.edu.hk,
zikang.wang@connect.polyu.hk, }\\
{\tt
shan.yuheng@u.nus.edu,
heyanchai@cuhk.edu.hk,}\\
{\tt
\{lijiayi03531, zxr15110781616\}@gmail.com,
liyu@cse.cuhk.edu.hk
}
}

\begin{document}
\maketitle
\begin{abstract}
Large Language models (LLMs) have emerged as powerful tools for addressing challenges across diverse domains. Notably, recent studies have demonstrated that large language models significantly enhance the efficiency of biomolecular analysis and synthesis, attracting widespread attention from academics and medicine. In this paper, we systematically investigate the application of prompt-based methods with LLMs to biological sequences, including DNA, RNA, proteins, and drug discovery tasks. Specifically, we focus on how prompt engineering enables LLMs to tackle domain-specific problems, such as promoter sequence prediction, protein structure modeling, and drug-target binding affinity prediction, often with limited labeled data. Furthermore, our discussion highlights the transformative potential of prompting in bioinformatics while addressing key challenges such as data scarcity, multimodal fusion, and computational resource limitations. Our aim is for this paper to function both as a foundational primer for newcomers and a catalyst for continued innovation within this dynamic field of study.

\end{abstract}

\section{Introduction}



\begin{figure}[!t]
	\setlength{\belowcaptionskip}{-8pt}
	\centering
	\includegraphics[width=1\linewidth]{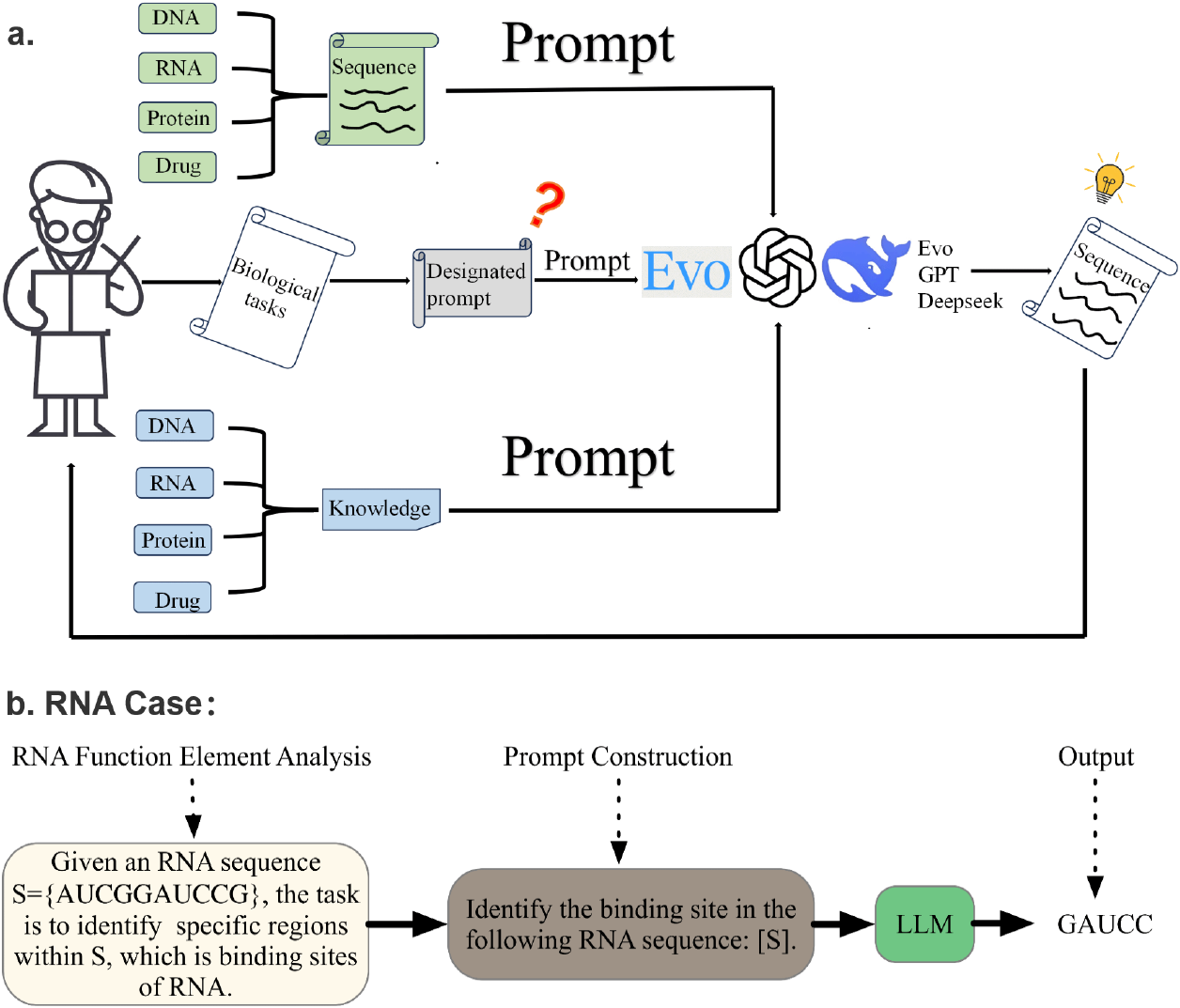}
	\caption{Biological sequence with language model prompting, and a RNA prompting case.} 
	\label{fig:example}
\end{figure}

LLMs have demonstrated remarkable advancements, primarily due to their capabilities in modeling the hidden relationships within textual sequences \cite{achiam2023gpt,dubey2024llama}. This innovation presents a unique opportunity for bioinformatics, where biological sequences (e.g., DNA, RNA, and proteins) exhibit structural similarities to natural languages. By leveraging LLMs, researchers can extract meaningful patterns from these sequences, enabling notable breakthroughs in diverse downstream tasks, represented by classification, structure prediction, and drug discovery. Despite their great potential, there still remain several challenges while applying LLMs to biological sequence analysis. One of the most pressing issues is data scarcity, as labeled biological datasets are always expensive and labor-intensive to obtain. This limitation hinders the effectiveness of supervised learning approaches. We give a specific example shown in Figure~\ref{fig:example}b.

\begin{figure*}[!htbp]
    \centering
    \includegraphics[width=1.0\textwidth]{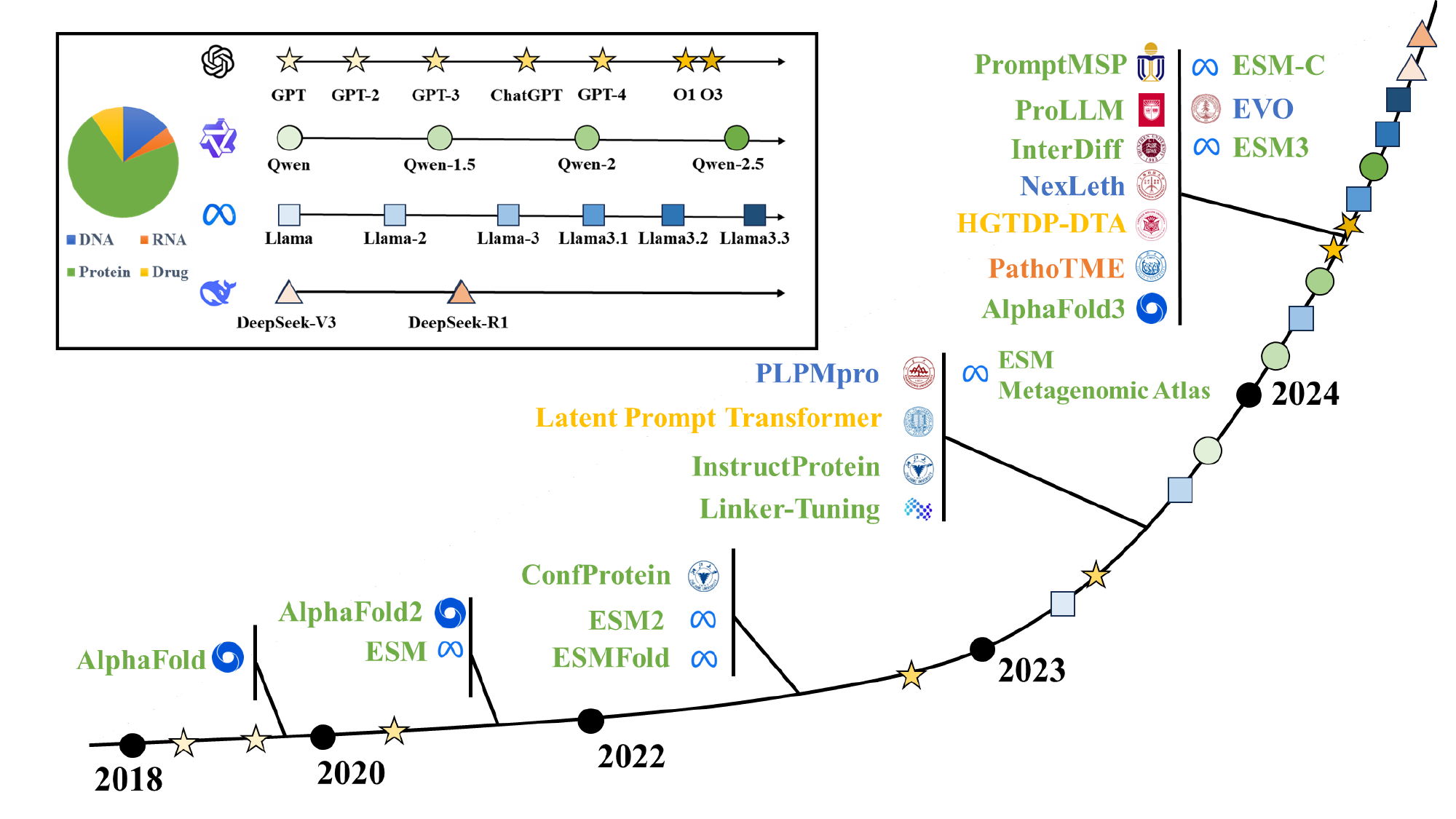}
    \caption{Timeline of key advancements in LLMs for computational biology.}
    \label{fig:overview}
\end{figure*}

To address this issue, prompt engineering has emerged as a powerful strategy that could enhance the adaptability of LLMs. Specifically, prompt-based methods leverage in-context learning, allowing LLMs to perform zero-shot and few-shot learning on biological tasks with minimal labeled data. Representative models (e.g., BERT~\cite{devlin2018bert}, GPT~\cite{achiam2023gpt}, ProtBERT~\cite{btac020} and ESM~\cite{lin2022language}) have been successfully adapted to biological sequences through precisely designed prompts, enabling them to generalize across diverse tasks represented by promoter sequence prediction, protein structure modeling, and drug-target binding affinity prediction. Overall, the timeline in Figure \ref{fig:overview} illustrates several key milestones and advancements in LLMs for computational biology.

\textbf{Organization of This Survey :} In this paper, we conduct the first survey of recent advancements in biological sequence with language model prompting. Specifically, we begin by introducing the fundamentals of biological sequences and explaining how prompt engineering facilitates LLMs applications in various downstream taks accross different domains (§\ref{2}). Specific examples of applications in various fields are also given (Figure \ref{fig:wetq}). We then present a detailed survey of prompting methods (§\ref{3}), categorizing existing approaches based on their applications in bioinformatics (Figure \ref{fig:lit_surv}). Meanwhile, we also examine the transformative role of AlphaFold and the Evolutionary Scale Modeling (ESM) series (§\ref{3}), highlighting their contributions to protein structure prediction and proteome-scale modeling. Next, we outline several key challenges (§\ref{4}) such as data scarcity and high labeling costs then explore future research directions (§\ref{5}), focusing on multi-modal prompt fusion, efficient adaptation techniques and data-centric annotation strategies. Finally, we conclude this survey (§\ref{6}) by summarizing key insights and underscoring the role of prompt engineering in advancing AI-driven biological research.

\section{Biological Sequence Prompting Tasks}
\label{2}
In this section, we first introduce the concept of biological sequences and how language modeling and prompt engineering fit into bioinformatics. Then, we outline how various biological tasks can be formulated as NLP problems and provide a concise categorization of their application domains (DNA, RNA, proteins, and drug discovery), setting the stage for the methodological discussions in subsequent sections.

\subsection{Biological Sequences}
Biological sequences, such as DNA, RNA, and proteins, can be viewed as linear arrangements of tokens from their respective alphabets. For instance, DNA and RNA use nucleotides \(\{\text{A}, \text{C}, \text{G}, \text{T/U}\}\), while proteins typically involve 20 standard amino acids. Formally, a biological sequence of length \(L\) is denoted as
\begin{equation}
S = \{x_1, \dots, x_L\},
\end{equation}
where each token \(x_i\) is drawn from an alphabet \(\mathcal{A}\). These sequences carry crucial information for cellular processes, such as gene expression, protein folding, and molecular interactions. Understanding and modeling these sequences is central to many tasks in computational biology.

\subsection{Prompt Engineering} 
Prompts further enhance the power of LLMs by integrating task-specific cues into the input of the model. This is achieved through textual templates or continuous embeddings, which we denote abstractly as:
\begin{equation}
T = f(S, P; \theta),
\end{equation}
where \(S\) represents biological sequence, \(P\) encodes domain-specific information as a prompt, and \(\theta\) denotes the model parameters. Prompting is particularly valuable in data-scarce scenarios (e.g., zero-shot or few-shot learning), guiding the model to focus on relevant biological patterns.

\subsection{Overview of Prompt-Based Task Mapping}
Biological tasks such as promoter prediction or protein-ligand binding can be effectively reinterpreted as NLP tasks using prompts. For instance, in promoter prediction, specific segments of a DNA sequence can be replaced with \texttt{[MASK]} tokens, transforming the task into a masked language modeling problem. Similarly, protein-ligand binding affinity prediction can be framed as a question-answering or fill-in-the-blank prompt, focusing on molecular compatibility. By carefully designing prompts, researchers leverage the extensive pretraining of LLMs to achieve strong performance even with limited labeled data. In this section, we provide an overview of four key application areas in biological sequence analysis:
\subsubsection{DNA}
 \textbf{Promoter Identification.} Detects promoter regions in DNA sequences, capturing key motifs such as the TATA-box to understand gene regulation. Benefiting from the remarkable performance of prompting, we effectively reinterpreted this task as an NLP task by constructing proper prompts, which can be formally defined as follows:
 \begin{equation}
     H = M_{\text{DNA}}(S, P_{\text{prom}}),
 \end{equation}
where $S$ is the input DNA sequence, $M_{\text{DNA}}$ is the large language model of DNA for promoter identification, $P_{\text{prom}}$ is is the constructed prompt, and $H$ is the predicted promoter region with functional motifs.

 For example, given a sequence $S=\{ATGCGATACTAGGATATAAGCTAG\}$. To detect the promoter region in this DNA sequence, we design this prompt: "Locate the promoter region between positions X-Y in [S] containing TATA-box motifs." Then the DNA sequence and constructed prompts are inputted into the language models to generate more accurate results.

\noindent\textbf{Mechanism Explanation.} Generates interpretable insights into synthetic lethality (SL) mechanisms, aiding in identifying potential cancer drug targets. Benefiting from the remarkable performance of prompting, we effectively reinterpreted this task as an NLP task by constructing proper prompts, which can be formally defined as follows:
\begin{equation}
    J = M_{\text{DNA}}(S_g, P_{\text{sl}}),
\end{equation}
where $S_g$ is the input gene pair, $M_{\text{DNA}}$ is the large language model of DNA for mechanism explanation, $P_{\text{sl}}$ is the constructed prompt, and $J$ is the structured explanation of synthetic lethal interactions.

 For instance, given a gene pair $S_g=\{BRCA1, PARP1\}$. To explain the SL mechanism, we construct a corresponding prompt: "Explain the synthetic lethality mechanism between [Gene A] and [Gene B], focusing on DNA repair pathways". Then, the gene pair and constructed prompts are inputted into the language models to generate more accurate results.

\subsubsection{RNA}
 \textbf{RNA Functional Element Analysis.} Identifies and characterizes splicing signals, regulatory elements, and sequence motifs associated with gene expression regulation. Benefiting from the remarkable performance of prompting, we effectively reinterpreted this task as an NLP task by constructing proper prompts, which can be formally defined as follows:
\begin{equation}
        R = M_{\text{RNA}}(S,P_f),
\end{equation}
where $S$ is the input RNA sequence, $M_{\text{RNA}}$ is the large language model of RNA for functional element analysis, $P_f$ is the constructed prompt, and $R$ is the generated response, which corresponds to the functional element $E$.

 Moreover, we take a specific example to illustrate how to use a prompt-based method to model RNA-related tasks as NLP tasks. For instance, given an RNA sequence $S=\{AUCGGAUCCG\}$, we want to identify specific regions within $S$, which are binding sites of RNA. We can construct a corresponding prompt: "Identify the binding site in the following RNA sequence: [$S$]". Then, the RNA sequence and constructed prompts are inputted into the language models to generate more accurate results.

\noindent\textbf{Cell Type Annotation.} Automates the classification of cell types in single-cell RNA sequencing (scRNA-seq) data, improving accuracy in diverse and noisy datasets. Benefiting from the remarkable performance of prompting, we effectively reinterpreted this task as an NLP task by constructing proper prompts, which can be formally defined as follows:
\begin{equation}
    L = M_{\text{RNA}}(S_c, P_{\text{cell}}),
\end{equation}
where $S_c$ is the input gene expression profile, $M_{\text{RNA}}$ is large language model of RNA for cell type annotation, $ P_{\text{cell}}$ is the constructed prompt, and $L$ corresponds to the predicted cell type labels with confidence scores. 

 Moreover, we take a specific example to illustrate how to use a prompt-based method to model RNA-related tasks as NLP tasks. For instance, given a cell expression profile $S_c=\{CD3E:12.8, CD8A:9.4, CD19:0.3, MS4A1:0.1\}$, we construct a corresponding prompt: "Classify this scRNA-seq cell's type based on top expressed genes: [S], and include canonical markers". Then, the cell expression profile and constructed prompts are inputted into the language models to generate more accurate results.

\subsubsection{Protein}
 \textbf{Protein Structure Modeling and Prediction.} Focuses on determining the three-dimensional structure of proteins from amino acid sequences, which is crucial for understanding function and interactions. Benefiting from the remarkable performance of prompting, we effectively reinterpreted this task as an NLP task by constructing proper prompts, which can be formally defined as follows:
\begin{equation}
    \Phi = M_{\text{protein}}(S, P_{\text{fold}})
\end{equation}
where $S$ is the input amino acid sequence, $M_{\text{protein}}$ is large language model of protein for structure modeling and prediction, $P_{\text{fold}}$ is the is the constructed prompt, and $\Phi$ corresponds to the predicted structural coordinates.

 For example, given a sequence $S=\{GVNPGVAPLSLLI\}$, we can construct a corresponding prompt: "Predict the tertiary structure topology for [S] with secondary structure annotations". Then, the protein sequence and constructed prompts are inputted into the language models to generate more accurate results.

\noindent\textbf{Molecular Interaction Modeling.} Studies how proteins interact with other molecules, including ligands and other proteins, to inform drug design and functional analysis.Benefiting from the remarkable performance of prompting, we effectively reinterpreted this task as an NLP task through constructing proper prompts, which can be formally defined as follows:
\begin{equation}
    \Gamma = M_{\text{protein}}(S_l, P_{\text{interact}}),
\end{equation}
where $S_l$ is the input molecular pair, $M_{\text{protein}}$ is large language model of protein for molecular interaction modeling, $P_{\text{interact}}$ is the is the constructed prompt, and $\Gamma$ corresponds to the predicted binding parameters. 

 For instance, given a kinase-ligand pair $S_l = \{EGFR kinase: MGPSV..., Gefitinib: C1=CN=CC=C1\}$, we can construct a corresponding prompt: "Predict binding mode between EGFR kinase and Gefitinib, identifying critical hydrogen bonds and hydrophobic contacts". Then, the molecular pair and constructed prompts are inputted into the language models to generate more accurate results.
 
\noindent\textbf{Protein Language-Based Generation.} Explores the generation of protein sequences and functional annotations using language-inspired models, facilitating de novo protein design. Benefiting from the remarkable performance of prompting, we effectively reinterpreted this task as an NLP task by constructing proper prompts, which can be formally defined as follows:
\begin{equation}
    \Psi = M_{\text{protein}}(P_{\text{design}}, \theta), 
\end{equation}
where $\theta$ is functional constraints, $M_{\text{protein}}$ is a large language model of protein for de novo protein design, $P_{\text{design}}$ is the constructed prompt, and $\Psi$ corresponds to the generated protein sequence with structural and functional metadata.

 For example, Given a target,to engineer a thermostable enzyme, we can construct a corresponding prompt: "Generate a $\beta$-lactamase variant with enhanced thermal stability and maintained catalytic efficiency". Then, the constructed prompts is inputted into the language models to generate more accurate results.
 
\noindent\textbf{Other Protein-Related Tasks.} Includes \emph{Polypeptide Design}, \emph{Conformation Perception}, and \emph{Protein Interaction Reasoning}, expanding applications in structural and functional biology. Benefiting from the remarkable performance of prompting, we effectively reinterpreted this task as an NLP task by constructing proper prompts, which can be formally defined as follows:
\begin{equation}  
    K = M_{\text{protein}}(S_r, P_{\text{task}}),  
\end{equation}  
where $S_r$ are the task-specific inputs, $M_{\text{protein}}$ is a large language model of protein for other tasks, $P_{\text{task}}$ is the constructed prompt, and $K$ corresponds to multimodal outputs spanning sequences, structures, and mechanistic insights.

 For easy understanding, we will illustrate how these protein-related tasks can be modeled as NLP tasks using a cue-based approach by following three concrete examples.
 First, in order to generate antimicrobial peptides, we can construct a corresponding prompt: "Design a 15-residue cationic $\alpha$-helical peptide targeting Gram-negative bacteria with <10\% hemolysis". Second, given a kinase sequence $S_r=\{IGPGRAFVT\}$, in order to predict conformational, we can construct a corresponding prompt: "Predict conformational changes upon ATP binding". Finally, given a ubiquitin-ligase pair $S_r=\{Ubiquitin, E6AP\}$. To reason about protein interactions, we can construct a corresponding prompt: "Infer recognition mechanism for Ub-E6AP complex formation". Then, the inputs and constructed prompts are inputted into the language models to generate more accurate results.
 
\subsubsection{Drug Discovery}
 \textbf{Drug-Target Binding Prediction.} Focuses on estimating the binding affinity between drugs and their molecular targets, aiding in the identification of potential therapeutics. Benefiting from the remarkable performance of prompting, we effectively reinterpreted this task as an NLP task by constructing proper prompts, which can be formally defined as follows:
 \begin{equation}  
    N = M_{\text{drug}}(S_d, S_t, P_{\text{bind}}),  
 \end{equation}
 where $S_d$ is the drug molecular structure, $S_t$ is the target protein sequence, $M_{\text{drug}}$ is the large language model of drug for target binding prediction, $P_{\text{bind}}$ is the constructed prompt, and N corresponds to predicted binding metrics.

For example, given the anticancer drug Gefitinib and EGFR kinase, we can construct a corresponding prompt: "Predict binding affinity and critical interactions between Gefitinib and EGFR kinase".Then, the inputs and constructed prompts are inputted into the language models to generate more accurate results.

\noindent\textbf{Molecular Design.} Involves generating and optimizing molecular structures to achieve desired pharmaceutical properties, such as bioavailability and synthetic accessibility. Benefiting from the remarkable performance of prompting, we effectively reinterpreted this task as an NLP task by constructing proper prompts, which can be formally defined as follows:
\begin{equation}  
    U = M_{\text{drug}}(\theta, P_{\text{design}}),  
\end{equation}  
where $\theta$ is target property specifications, $M_{\text{drug}}$ is a large language model of drug for molecular design, $P_{\text{design}}$ is the constructed prompt, and U corresponds to the generated molecule. 

 For instance, to design a non-steroidal anti-inflammatory drug (NSAID) with reduced gastrointestinal toxicity, we can construct a corresponding prompt: "Generate a COX-2 selective inhibitor with $IC_{50} < 10 nM$, LogP 2.5–3.5, and $>80\%$ plasma stability at 24h". Then, the constructed prompts are inputted into the language models to generate more accurate results.

In the next sections, we will delve deeper into the specific methods, discussing their performance and how prompt-based strategies are reshaping biological sequence analysis. By encoding domain knowledge through carefully designed prompts, LLMs have become powerful tools for tackling complex tasks in DNA, RNA, protein, and drug discovery.
\begin{figure*}[h!]
    \centering
    \includegraphics[width=1.0\textwidth]{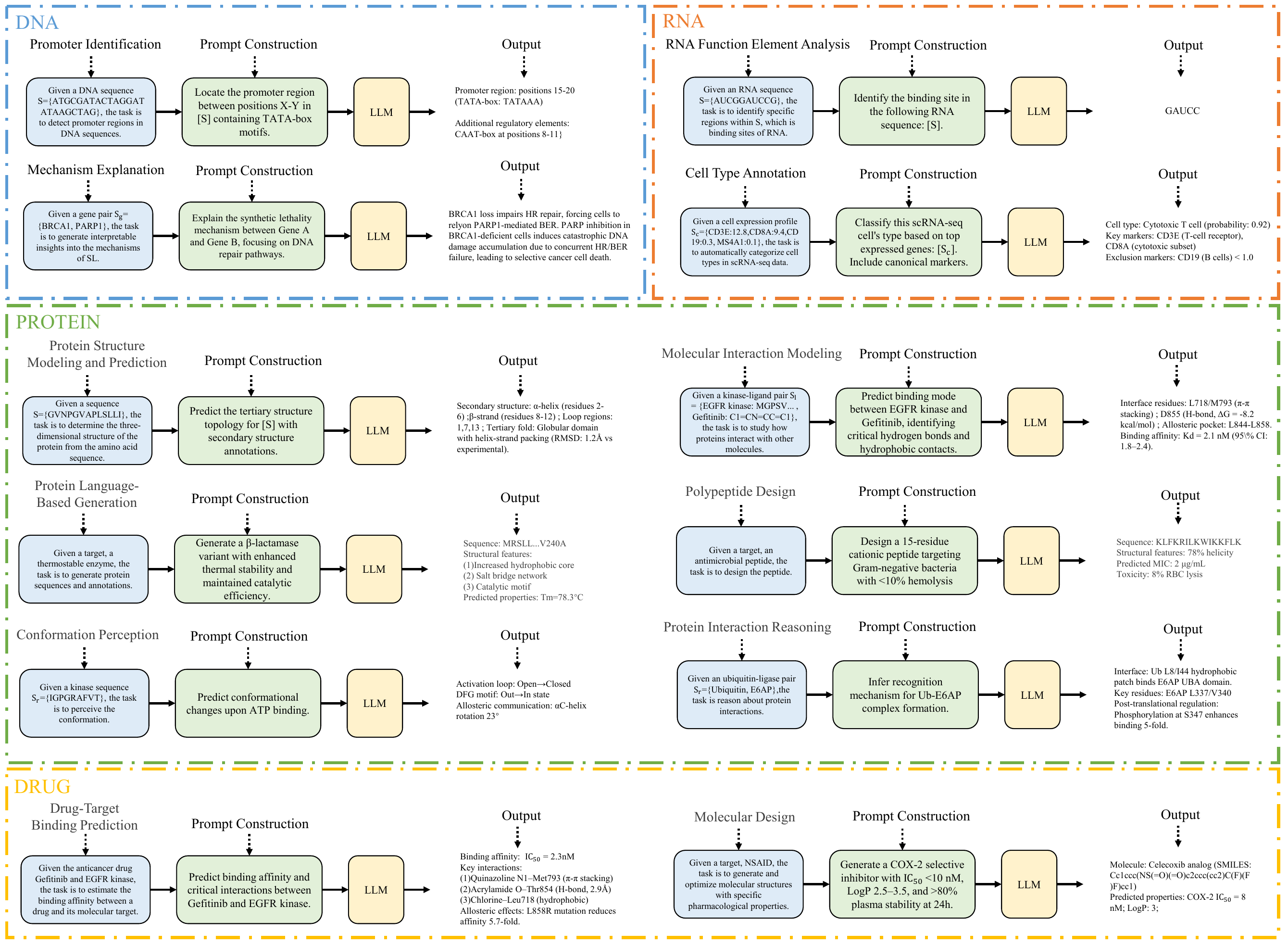}
    \caption{Specific examples of the use of prompt methods in DNA, RNA, protein and drug.}
    \label{fig:wetq}
\end{figure*}

\section{Prompting Applications in Biological Sequences}
\label{3}
Prompting technology, which is especially suitable for zero/few sample learning by designing prompts for pre-trained models to guide them to accomplish specific tasks, has been widely applied across various domains. It is extensively used in natural language processing (e.g., GPT, LLaMA for text categorization, summarization, Q\&A, etc.) and bioinformatics (e.g., DNABERT for recognizing DNA sequences, GPT-3/4 for annotating RNA-seq data, and sequential prompts for improving protein structure prediction and molecular design). By embedding task-relevant information in prompts, this technique focuses the model's attention on producing high-quality categorization, prediction, or generative output, making it an effective tool for solving complex problems due to its flexibility and data efficiency. The following sections categorize and summarize prompting methods across DNA, RNA, protein, and drug discovery domains. The detailed taxonomy is shown in Figure~\ref{fig:lit_surv}. In addition, Figure~\ref{fig:wetq} is figure of specific examples of the use of prompt methods in DNA, RNA, protein, and drug.

\subsection{DNA Sequences}

\noindent\textbf{NexLeth} \cite{zhang2024prompt} is a novel approach for generating natural language explanations of synthetic lethality (SL) mechanisms, which are critical for cancer drug discovery. The NexLeth pipeline integrates SL knowledge graphs with personalized prompt templates, enhancing explainability and textual coherence in SL predictions.

\noindent\textbf{PLPMpro} \cite{li2023plpmpro} is a prompt-learning framework that leverages pre-trained models such as DNABERT for promoter sequence prediction. By employing soft templates and verbalizers, PLPMpro effectively captures biologically meaningful sequence patterns, such as the TATA-box motif, achieving state-of-the-art performance.

\subsection{RNA Sequences}

\noindent\textbf{PathoTME} \cite{meng2024genomicsguided} is a genomics-guided deep learning framework for tumor microenvironment (TME) subtype prediction. Utilizing visual prompt tuning (VPT) and domain adversarial networks, PathoTME achieves superior classification performance while addressing tissue heterogeneity challenges.

\noindent\textbf{GPTCellType} \cite{hou2024assessing} applies GPT-4 for automated cell type annotation in single-cell RNA sequencing (scRNA-seq). GPTCellType outperforms traditional methods, demonstrating robustness to noisy datasets and underscoring the potential of large language models (LLMs) for RNA-related analyses.

\subsection{Protein Sequences}
\noindent\textbf{InterDiff} \cite{wu2024guided} is a diffusion-based molecular generative model. By incorporating interaction prompts, InterDiff guides molecular design for protein-ligand interactions and achieves superior performance in predicting binding affinity and interaction specificity.

\noindent\textbf{Linker-Tuning} \cite{zou2023linker} is a lightweight adaptation method for large protein language models such as ESMFold. This approach improves the prediction of heterodimeric protein structures and achieves competitive accuracy while reducing computational costs.

\noindent\textbf{InstructProtein} \cite{wang2023instructprotein} is a bidirectional framework that bridges protein and human languages. By leveraging instruction tuning, this model excels at zero-shot protein function annotation and de novo sequence design.

\noindent\textbf{PromptMSP} \cite{gao2024proteinmultimerstructureprediction} enhances multimer structure prediction through just-in-time learning and meta-learning.PromptMSP integrates conditional PPI knowledge and improves accuracy by reorganizing multimer prediction into fixed-scale tasks. The results for PDB-M show that the method outperforms the baseline method.

\noindent\textbf{ConfProtein} \cite{zhang2022promptguidedinjectionconformationpretrained} enhances pre-trained protein models (PTPMs) with the help of prompt learning, integrating sequence and interaction conformational cues to capture protein conformations. From the results, ConfProtein improves PPI prediction and antibody binding while maintaining sequence correlation performance, and is effectively validated on multiple datasets.

\noindent\textbf{ProLLM} \cite{jin2024prollmproteinchainofthoughtsenhanced} leverages LLMs and the Protein Chain of Thought (ProCoT) mechanism to model direct and indirect protein-protein interactions (PPIs) as inference tasks. By integrating ProtTrans embeddings and instruction fine-tuning, it achieves state-of-the-art performance in PPI prediction.

\subsection{Drug Discovery}
\noindent\textbf{HGTDP-DTA} \cite{xiao2024hgtdp} is a hybrid Graph-Transformer framework with dynamic prompt generation for drug-target binding affinity prediction. This model integrates both graph-based and sequence-based representations, achieving superior performance over state-of-the-art methods on benchmark datasets.

\noindent\textbf{Latent Prompt Transformer} \cite{kong2024dual} is a generative framework for molecule design. By incorporating latent prompts into a unified architecture, the Latent Prompt Transformer achieves state-of-the-art performance in multi-objective molecule optimization and drug-like molecule generation.

\noindent\textbf{In-Context Learning for Drug Synergy Prediction} \cite{edwards2023synergpt} introduces an in-context learning strategy for predicting synergistic drug combinations. By leveraging masking techniques and graph representations, this approach enhances personalized drug synergy prediction.

\subsection{The Significance of AlphaFold and ESM}
\noindent\textbf{AlphaFold.}
AlphaFold has revolutionized protein structure prediction, providing near-atomic accuracy in determining 3D structures directly from amino acid sequences. By leveraging attention-based architectures, it has resolved the long-standing challenge of protein folding, significantly advancing fields such as drug discovery, enzyme engineering, and disease modeling.

\noindent\textbf{Evolutionary Scale Modeling (ESM) Series.}
The ESM series, including ESMFold and ESM-2, represents a major leap in protein language modeling by enabling high-throughput structure prediction without the need for multiple sequence alignments (MSAs). ESM-3 further expands this capability, integrating sequence, structure, and function while maintaining computational efficiency. These models are invaluable for large-scale proteome analysis, facilitating the study of poorly characterized protein families.

\section{Challenges}
\label{4}
\noindent\textbf{Data Scarcity and High Labeling Costs.}
High-quality datasets for DNA/RNA sequences, such as NexLeth for synthetic lethal (SL) gene pairs, often require labor-intensive literature mining and expert review, making them expensive and limited in scale. This issue is particularly pronounced for rare diseases and minority species, where datasets are extremely scarce~\cite{graefe2025ontology}. In protein studies, the number of experimentally determined structures and reliable annotations remains insufficient, especially for multi-chain complexes, protein-protein interactions, and diverse conformations. Without robust experimental validation or high-confidence labels, the performance improvements achieved through prompt-based methods are often difficult to replicate reliably~\cite{chen2024applying}. Similarly, in drug discovery, drug-target affinity (DTA) data is both scarce and costly to obtain. This limits the ability of prompt-based methods to generalize effectively across small samples, different species, or novel targets~\cite{pei2023breaking}.

\noindent\textbf{Difficulties in Multimodal Feature Fusion.}
Bioinformatics often requires simultaneous processing of sequence, structure, image, and phenotypic data. While methods like PathoTME have explored combining visual prompts with genomic information for tumor subtype prediction, effectively fusing high-dimensional data—such as images, protein 3D structures, transcriptomes, and molecular graphs—under a unified prompt-based framework remains a significant challenge~\cite{koh2024physicochemical}. For more complex scenarios, such as protein-ligand interactions or polymer structures, the contextual and dependency information required for effective prompt encoding becomes even more extensive. However, the scalability of existing models to handle such complexity remains limited~\cite{cao2024surfdock}.

\noindent\textbf{Computational Resource Constraints.}
Large-scale pre-trained models, such as ESM-2 and Evo~\cite{zhang2024protein,nguyen2024sequence}, excel at handling long sequences and large datasets but come with substantial computational costs and training overhead. While higher-order models like AlphaFold-3 and ESMFold have achieved remarkable accuracy improvements, their inference speed and hardware requirements pose significant barriers for academic or small-to-medium research institutions. Additionally, prompt-based approaches often require task-specific designs, such as linker-tuning or dynamic prompt generation, to adapt to different downstream tasks or data distributions~\cite{giray2023prompt}. Without efficient strategies, the cost of model iteration and optimization can become prohibitively high~\cite{ye2022unreliability}.

\section{Future directions}
\label{5}
\noindent\textbf{Data-Centric Synthetic Annotation Methods.}
To cope with the problem of data scarcity, future work should approach like exploring semi-supervised learning, domain adaptation, and synthetic data generation~\cite{zha2025data,hu2024reinforcement,zhang2024dynamic}. For example, using generative models to augment limited experimental samples. Also, using active learning frameworks to more effectively go about guiding the annotation process and help prompt-based models generalize even in resource-poor domains~\cite{zhao2020simplifying}.

\begin{figure}[!t]
	\setlength{\belowcaptionskip}{-8pt}
	\centering
	\includegraphics[width=1\linewidth]{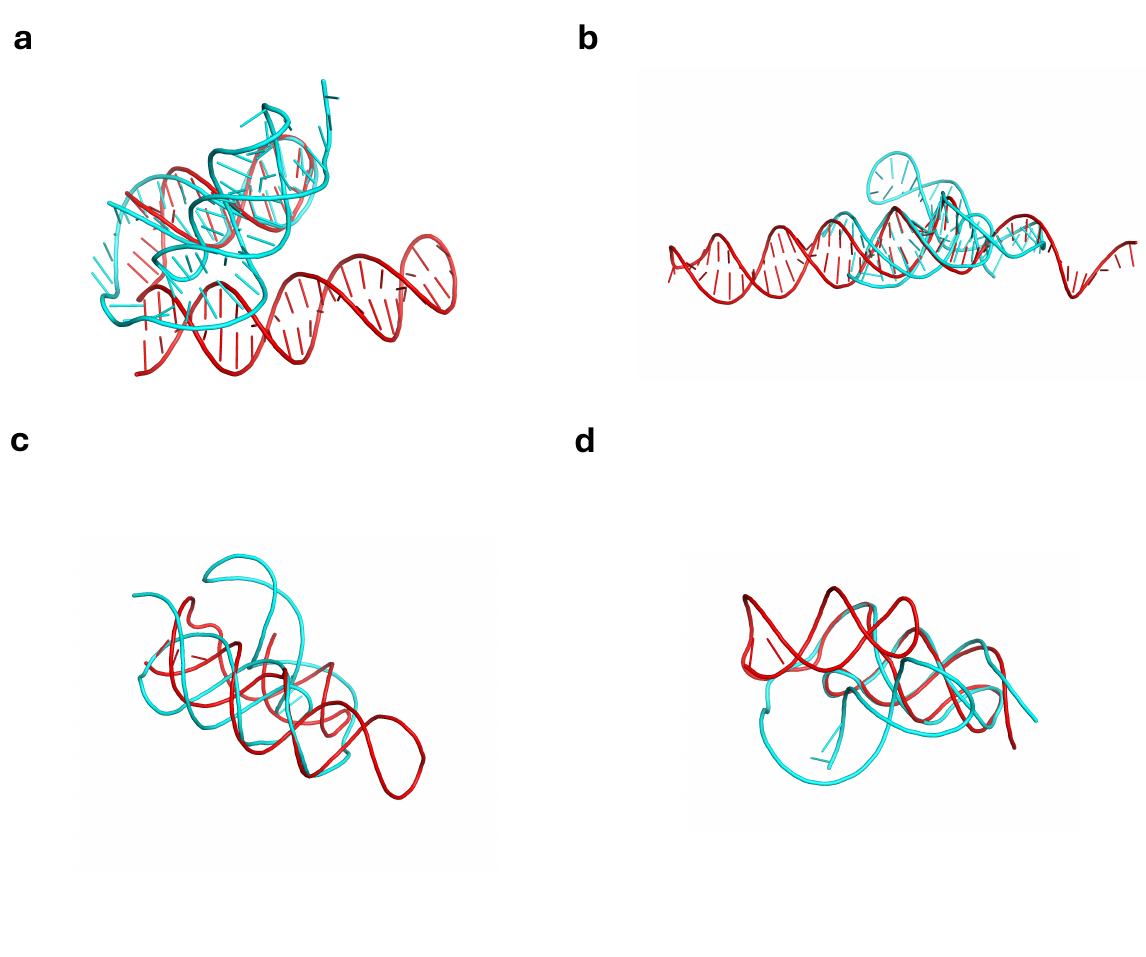}
	\caption{Case study on the prompt-based generated DNA sequences related to Landau Kleffner Syndrome (LKS). (a) DeepSeek-R1 (RMSD = 21.87 \AA, Tm-score = 0.20958) (b) GPT-4o (RMSD = 33.72 \AA, Tm-score = 0.14644) (c) Llama-3.3-70b (RMSD = 23.73 \AA, Tm-score = 0.14224) (d) Qwen-2.5-max (RMSD = 9.58 \AA, Tm-score = 0.44619).} 
	\label{fig:case}
\end{figure}

\noindent\textbf{Multi-Modal Prompt Fusion.}
In addition to sequence-level prompts, unifying structural, image, and metagenomic data, among others, is a future direction~\cite{liu2024deepseek}. Meanwhile, designing consistent cross-modal prompts and specialized attention layers helps models capture more complex correlations~\cite{ampazis2024diversifying}. In addition to this, increasing the interpretability of visualization modules will further improve trust and usability~\cite{munia2025attention}. Herein, we also examine the performances of four different LLMs in generating DNA sequences related to Landau Kleffner Syndrome (LKS), a rare brain condition that causes children to lose their ability to speak and understand language, after typing a personalized prompting (Figure \ref{fig:case} ).

\noindent\textbf{Lightweight and Efficient Adaptation.}
To address computational resource constraints, the use of techniques such as quantization, model pruning~\cite{cheng2024survey}, knowledge refinement~\cite{subagdja2024machine}, and low-rank adaptation (LoRA)~\cite{hu2021lora, wang2024prolora, wang2024mos} can reduce the size of parameters while maintaining predictive performance. Proper utilization of these prompting methods will make it easier for small labs with limited GPU resources to use large cue-based models, while also speeding up iterations in the model refinement process.


\section{Conclusion}
\label{6}
In this survey, we systematically reviewed the integration of prompt-based methods with LLMs for biological sequence analysis, spanning DNA, RNA, protein, and drug discovery. Our discussion demonstrated how prompt engineering enables LLMs to generalize across biological tasks, particularly in low-resource settings, by leveraging zero-shot and few-shot learning. Moving forward, we envision three promising research directions: (1) Data-Centric Prompt, (2) Unified Multimodal Prompting, and (3) Efficient and Scalable Prompting.

As LLM continues to evolve, prompt-based methodologies will play a crucial role in precision medicine and drug discovery. By addressing existing challenges and advancing these research directions, prompt engineering stands at the frontier of computational biology, unlocking new chances for the next generation of AI-driven bioinformatics.

\section*{Limitations}
This study is the first survey of recent advancements in biological sequence with language model prompting. We have made our best effort, but some limitations remain. We present recent methods and application domains rather than an exhaustive coverage. Due to space constraints, we can only provide brief method summaries without exhaustive technical details. Due to focusing primarily on publication from bioinformatics-related journals or conferences, we may have overlooked significant work published in other venues. We will continue to monitor the research community, incorporate new perspectives, and address any omissions in future updates.


\section*{Ethics Statement}

This paper does not involve ethics-related issues.

\bibliography{custom}

\begin{thebibliography}{41}
\providecommand{\natexlab}[1]{#1}

\bibitem[{Abramson et~al.(2024)Abramson, Adler, Dunger, Evans, Green, Pritzel, Ronneberger, Willmore, Ballard, Bambrick et~al.}]{abramson2024accurate}
Josh Abramson, Jonas Adler, Jack Dunger, Richard Evans, Tim Green, Alexander Pritzel, Olaf Ronneberger, Lindsay Willmore, Andrew~J Ballard, Joshua Bambrick, et~al. 2024.
\newblock Accurate structure prediction of biomolecular interactions with alphafold 3.
\newblock \emph{Nature}, pages 1--3.

\bibitem[{Achiam et~al.(2023)Achiam, Adler, Agarwal, Ahmad, Akkaya, Aleman, Almeida, Altenschmidt, Altman, Anadkat et~al.}]{achiam2023gpt}
Josh Achiam, Steven Adler, Sandhini Agarwal, Lama Ahmad, Ilge Akkaya, Florencia~Leoni Aleman, Diogo Almeida, Janko Altenschmidt, Sam Altman, Shyamal Anadkat, et~al. 2023.
\newblock Gpt-4 technical report.
\newblock \emph{arXiv preprint arXiv:2303.08774}.

\bibitem[{Ampazis and Sakketou(2024)}]{ampazis2024diversifying}
Nicholas Ampazis and Flora Sakketou. 2024.
\newblock Diversifying multi-head attention in the transformer model.
\newblock \emph{Machine Learning and Knowledge Extraction}, 6(4):2618--2638.

\bibitem[{Brandes et~al.(2022)Brandes, Ofer, Peleg, Rappoport, and Linial}]{btac020}
Nadav Brandes, Dan Ofer, Yam Peleg, Nadav Rappoport, and Michal Linial. 2022.
\newblock \href {https://doi.org/10.1093/bioinformatics/btac020} {Proteinbert: a universal deep-learning model of protein sequence and function}.
\newblock \emph{Bioinformatics}, 38(8):2102--2110.

\bibitem[{Cao et~al.(2024)Cao, Chen, Zhang, Wang, Huang, Yu, Jiang, Fan, Zhang, Zhou et~al.}]{cao2024surfdock}
Duanhua Cao, Mingan Chen, Runze Zhang, Zhaokun Wang, Manlin Huang, Jie Yu, Xinyu Jiang, Zhehuan Fan, Wei Zhang, Hao Zhou, et~al. 2024.
\newblock Surfdock is a surface-informed diffusion generative model for reliable and accurate protein--ligand complex prediction.
\newblock \emph{Nature Methods}, pages 1--13.

\bibitem[{Chen et~al.(2024)Chen, Yang, Cui, Kim, Talwalkar, and Ma}]{chen2024applying}
Valerie Chen, Muyu Yang, Wenbo Cui, Joon~Sik Kim, Ameet Talwalkar, and Jian Ma. 2024.
\newblock Applying interpretable machine learning in computational biology—pitfalls, recommendations and opportunities for new developments.
\newblock \emph{Nature methods}, 21(8):1454--1461.

\bibitem[{Cheng et~al.(2024)Cheng, Zhang, and Shi}]{cheng2024survey}
Hongrong Cheng, Miao Zhang, and Javen~Qinfeng Shi. 2024.
\newblock A survey on deep neural network pruning: Taxonomy, comparison, analysis, and recommendations.
\newblock \emph{IEEE Transactions on Pattern Analysis and Machine Intelligence}.

\bibitem[{Devlin(2018)}]{devlin2018bert}
Jacob Devlin. 2018.
\newblock Bert: Pre-training of deep bidirectional transformers for language understanding.
\newblock \emph{arXiv preprint arXiv:1810.04805}.

\bibitem[{Dubey et~al.(2024)Dubey, Jauhri, Pandey, Kadian, Al-Dahle, Letman, Mathur, Schelten, Yang, Fan et~al.}]{dubey2024llama}
Abhimanyu Dubey, Abhinav Jauhri, Abhinav Pandey, Abhishek Kadian, Ahmad Al-Dahle, Aiesha Letman, Akhil Mathur, Alan Schelten, Amy Yang, Angela Fan, et~al. 2024.
\newblock The llama 3 herd of models.
\newblock \emph{arXiv preprint arXiv:2407.21783}.

\bibitem[{Edwards et~al.(2023)Edwards, Naik, Khot, Burke, Ji, and Hope}]{edwards2023synergpt}
Carl Edwards, Aakanksha Naik, Tushar Khot, Martin Burke, Heng Ji, and Tom Hope. 2023.
\newblock Synergpt: In-context learning for personalized drug synergy prediction and drug design.
\newblock \emph{arXiv preprint arXiv:2307.11694}.

\bibitem[{Gao et~al.(2024)Gao, Sun, Liu, Li, Cheng, and Li}]{gao2024proteinmultimerstructureprediction}
Ziqi Gao, Xiangguo Sun, Zijing Liu, Yu~Li, Hong Cheng, and Jia Li. 2024.
\newblock \href {https://arxiv.org/abs/2402.18813} {Protein multimer structure prediction via prompt learning}.
\newblock \emph{Preprint}, arXiv:2402.18813.

\bibitem[{Giray(2023)}]{giray2023prompt}
Louie Giray. 2023.
\newblock Prompt engineering with chatgpt: a guide for academic writers.
\newblock \emph{Annals of biomedical engineering}, 51(12):2629--2633.

\bibitem[{Graefe et~al.(2025)Graefe, H{\"u}bner, Rehburg, Sander, Klopfenstein, Alkarkoukly, Gr{\"o}nke, Weyersberg, Danis, Zsch{\"u}ntzsch et~al.}]{graefe2025ontology}
Adam~SL Graefe, Miriam~R H{\"u}bner, Filip Rehburg, Steffen Sander, Sophie~AI Klopfenstein, Samer Alkarkoukly, Ana Gr{\"o}nke, Annic Weyersberg, Daniel Danis, Jana Zsch{\"u}ntzsch, et~al. 2025.
\newblock An ontology-based rare disease common data model harmonising international registries, fhir, and phenopackets.
\newblock \emph{Scientific Data}, 12(1):234.

\bibitem[{Hayes et~al.(2025)Hayes, Rao, Akin, Sofroniew, Oktay, Lin, Verkuil, Tran, Deaton, Wiggert et~al.}]{hayes2025simulating}
Thomas Hayes, Roshan Rao, Halil Akin, Nicholas~J Sofroniew, Deniz Oktay, Zeming Lin, Robert Verkuil, Vincent~Q Tran, Jonathan Deaton, Marius Wiggert, et~al. 2025.
\newblock Simulating 500 million years of evolution with a language model.
\newblock \emph{Science}, page eads0018.

\bibitem[{Hou and Ji(2024)}]{hou2024assessing}
Wenpin Hou and Zhicheng Ji. 2024.
\newblock Assessing gpt-4 for cell type annotation in single-cell rna-seq analysis.
\newblock \emph{Nature Methods}, pages 1--4.

\bibitem[{Hu et~al.(2021)Hu, Shen, Wallis, Allen-Zhu, Li, Wang, Wang, and Chen}]{hu2021lora}
Edward~J Hu, Yelong Shen, Phillip Wallis, Zeyuan Allen-Zhu, Yuanzhi Li, Shean Wang, Lu~Wang, and Weizhu Chen. 2021.
\newblock Lora: Low-rank adaptation of large language models.
\newblock \emph{arXiv preprint arXiv:2106.09685}.

\bibitem[{Hu et~al.(2024)Hu, Wang, Ying, and Fu}]{hu2024reinforcement}
Xuanming Hu, Dongjie Wang, Wangyang Ying, and Yanjie Fu. 2024.
\newblock Reinforcement feature transformation for polymer property performance prediction.
\newblock In \emph{Proceedings of the 33rd ACM International Conference on Information and Knowledge Management}, pages 4538--4545.

\bibitem[{Jin et~al.(2024)Jin, Xue, Wang, Kang, Ye, Zhou, Du, and Zhang}]{jin2024prollmproteinchainofthoughtsenhanced}
Mingyu Jin, Haochen Xue, Zhenting Wang, Boming Kang, Ruosong Ye, Kaixiong Zhou, Mengnan Du, and Yongfeng Zhang. 2024.
\newblock \href {https://arxiv.org/abs/2405.06649} {Prollm: Protein chain-of-thoughts enhanced llm for protein-protein interaction prediction}.
\newblock \emph{Preprint}, arXiv:2405.06649.

\bibitem[{Koh et~al.(2024)Koh, Nguyen, Pan, May, and Webb}]{koh2024physicochemical}
Huan~Yee Koh, Anh~TN Nguyen, Shirui Pan, Lauren~T May, and Geoffrey~I Webb. 2024.
\newblock Physicochemical graph neural network for learning protein--ligand interaction fingerprints from sequence data.
\newblock \emph{Nature Machine Intelligence}, pages 1--15.

\bibitem[{Kong et~al.(2024)Kong, Huang, Xie, Honig, Xu, Xue, Lin, Zhou, Zhong, Zheng et~al.}]{kong2024dual}
Deqian Kong, Yuhao Huang, Jianwen Xie, Edouardo Honig, Ming Xu, Shuanghong Xue, Pei Lin, Sanping Zhou, Sheng Zhong, Nanning Zheng, et~al. 2024.
\newblock Dual-space optimization: Improved molecule sequence design by latent prompt transformer.
\newblock \emph{arXiv preprint arXiv:2402.17179}.

\bibitem[{Li et~al.(2023)Li, Jin, Long, and Wei}]{li2023plpmpro}
Zhongshen Li, Junru Jin, Wentao Long, and Leyi Wei. 2023.
\newblock Plpmpro: Enhancing promoter sequence prediction with prompt-learning based pre-trained language model.
\newblock \emph{Computers in Biology and Medicine}, 164:107260.

\bibitem[{Lin et~al.(2022)Lin, Akin, Rao, Hie, Zhu, Lu, Smetanin, dos Santos~Costa, Fazel-Zarandi, Sercu, Candido et~al.}]{lin2022language}
Zeming Lin, Halil Akin, Roshan Rao, Brian Hie, Zhongkai Zhu, Wenting Lu, Nikita Smetanin, Allan dos Santos~Costa, Maryam Fazel-Zarandi, Tom Sercu, Sal Candido, et~al. 2022.
\newblock Language models of protein sequences at the scale of evolution enable accurate structure prediction.
\newblock \emph{bioRxiv}.

\bibitem[{Liu et~al.(2024)Liu, Feng, Xue, Wang, Wu, Lu, Zhao, Deng, Zhang, Ruan et~al.}]{liu2024deepseek}
Aixin Liu, Bei Feng, Bing Xue, Bingxuan Wang, Bochao Wu, Chengda Lu, Chenggang Zhao, Chengqi Deng, Chenyu Zhang, Chong Ruan, et~al. 2024.
\newblock Deepseek-v3 technical report.
\newblock \emph{arXiv preprint arXiv:2412.19437}.

\bibitem[{Meng et~al.(2024)Meng, Zhang, Yan, Chuai, Li, and Liu}]{meng2024genomicsguided}
Fangliangzi Meng, Hongrun Zhang, Ruodan Yan, Guohui Chuai, Chao Li, and Qi~Liu. 2024.
\newblock \href {https://arxiv.org/abs/2406.06517} {Genomics-guided representation learning for pathologic pan-cancer tumor microenvironment subtype prediction}.
\newblock \emph{Preprint}, arXiv:2406.06517.

\bibitem[{Munia et~al.(2025)Munia, Abdar, Hasan, Jalali, Banerjee, Khosravi, Hossain, Fu, and Frangi}]{munia2025attention}
Afsana~Ahmed Munia, Moloud Abdar, Mehedi Hasan, Mohammad~S Jalali, Biplab Banerjee, Abbas Khosravi, Ibrahim Hossain, Huazhu Fu, and Alejandro~F Frangi. 2025.
\newblock Attention-guided hierarchical fusion u-net for uncertainty-driven medical image segmentation.
\newblock \emph{Information Fusion}, 115:102719.

\bibitem[{Nguyen et~al.(2024)Nguyen, Poli, Durrant, Kang, Katrekar, Li, Bartie, Thomas, King, Brixi et~al.}]{nguyen2024sequence}
Eric Nguyen, Michael Poli, Matthew~G Durrant, Brian Kang, Dhruva Katrekar, David~B Li, Liam~J Bartie, Armin~W Thomas, Samuel~H King, Garyk Brixi, et~al. 2024.
\newblock Sequence modeling and design from molecular to genome scale with evo.
\newblock \emph{Science}, 386(6723):eado9336.

\bibitem[{Pei et~al.(2023)Pei, Wu, Zhu, Xia, Xie, Qin, Liu, Liu, and Yan}]{pei2023breaking}
Qizhi Pei, Lijun Wu, Jinhua Zhu, Yingce Xia, Shufang Xie, Tao Qin, Haiguang Liu, Tie-Yan Liu, and Rui Yan. 2023.
\newblock Breaking the barriers of data scarcity in drug--target affinity prediction.
\newblock \emph{Briefings in Bioinformatics}, 24(6):bbad386.

\bibitem[{Subagdja et~al.(2024)Subagdja, Shanthoshigaa, Wang, and Tan}]{subagdja2024machine}
Budhitama Subagdja, D~Shanthoshigaa, Zhaoxia Wang, and Ah-Hwee Tan. 2024.
\newblock Machine learning for refining knowledge graphs: A survey.
\newblock \emph{ACM Computing Surveys}, 56(6):1--38.

\bibitem[{Wang et~al.(2024{\natexlab{a}})Wang, Chen, Chen, Dong, Xue, Jiang, Kong, and Wu}]{wang2024mos}
Sheng Wang, Liheng Chen, Pengan Chen, Jingwei Dong, Boyang Xue, Jiyue Jiang, Lingpeng Kong, and Chuan Wu. 2024{\natexlab{a}}.
\newblock Mos: Unleashing parameter efficiency of low-rank adaptation with mixture of shards.
\newblock \emph{arXiv preprint arXiv:2410.00938}.

\bibitem[{Wang et~al.(2024{\natexlab{b}})Wang, Xue, Ye, Jiang, Chen, Kong, and Wu}]{wang2024prolora}
Sheng Wang, Boyang Xue, Jiacheng Ye, Jiyue Jiang, Liheng Chen, Lingpeng Kong, and Chuan Wu. 2024{\natexlab{b}}.
\newblock Prolora: Partial rotation empowers more parameter-efficient lora.
\newblock \emph{arXiv preprint arXiv:2402.16902}.

\bibitem[{Wang et~al.(2023)Wang, Zhang, Ding, Qin, Zhuang, Li, and Chen}]{wang2023instructprotein}
Zeyuan Wang, Qiang Zhang, Keyan Ding, Ming Qin, Xiang Zhuang, Xiaotong Li, and Huajun Chen. 2023.
\newblock Instructprotein: Aligning human and protein language via knowledge instruction.
\newblock \emph{arXiv preprint arXiv:2310.03269}.

\bibitem[{Wu et~al.(2024)Wu, Du, Yan, Lee, Bai, and Wu}]{wu2024guided}
Peng Wu, Huabin Du, Yingchao Yan, Tzong-Yi Lee, Chen Bai, and Song Wu. 2024.
\newblock Guided diffusion for molecular generation with interaction prompt.
\newblock \emph{Briefings in Bioinformatics}, 25(3):bbae174.

\bibitem[{Xiao et~al.(2024)Xiao, Wang, Xie, Zhu, Chen, Li, Wang, and Xu}]{xiao2024hgtdp}
Xi~Xiao, Wentao Wang, Jiacheng Xie, Lijing Zhu, Gaofei Chen, Zhengji Li, Tianyang Wang, and Min Xu. 2024.
\newblock Hgtdp-dta: Hybrid graph-transformer with dynamic prompt for drug-target binding affinity prediction.
\newblock \emph{arXiv preprint arXiv:2406.17697}.

\bibitem[{Ye and Durrett(2022)}]{ye2022unreliability}
Xi~Ye and Greg Durrett. 2022.
\newblock The unreliability of explanations in few-shot prompting for textual reasoning.
\newblock \emph{Advances in neural information processing systems}, 35:30378--30392.

\bibitem[{Zha et~al.(2025)Zha, Bhat, Lai, Yang, Jiang, Zhong, and Hu}]{zha2025data}
Daochen Zha, Zaid~Pervaiz Bhat, Kwei-Herng Lai, Fan Yang, Zhimeng Jiang, Shaochen Zhong, and Xia Hu. 2025.
\newblock Data-centric artificial intelligence: A survey.
\newblock \emph{ACM Computing Surveys}, 57(5):1--42.

\bibitem[{Zhang et~al.(2024{\natexlab{a}})Zhang, Feng, and Zheng}]{zhang2024prompt}
Ke~Zhang, Yimiao Feng, and Jie Zheng. 2024{\natexlab{a}}.
\newblock Prompt-based generation of natural language explanations of synthetic lethality for cancer drug discovery.
\newblock In \emph{Proceedings of the 2024 Joint International Conference on Computational Linguistics, Language Resources and Evaluation (LREC-COLING 2024)}, pages 13131--13142.

\bibitem[{Zhang et~al.(2022)Zhang, Wang, Han, Yu, Jin, and Chen}]{zhang2022promptguidedinjectionconformationpretrained}
Qiang Zhang, Zeyuan Wang, Yuqiang Han, Haoran Yu, Xurui Jin, and Huajun Chen. 2022.
\newblock \href {https://arxiv.org/abs/2202.02944} {Prompt-guided injection of conformation to pre-trained protein model}.
\newblock \emph{Preprint}, arXiv:2202.02944.

\bibitem[{Zhang et~al.(2024{\natexlab{b}})Zhang, Zhang, Rekabdar, Zhou, Wang, and Liu}]{zhang2024dynamic}
Xinhao Zhang, Jinghan Zhang, Banafsheh Rekabdar, Yuanchun Zhou, Pengfei Wang, and Kunpeng Liu. 2024{\natexlab{b}}.
\newblock Dynamic and adaptive feature generation with llm.
\newblock \emph{arXiv preprint arXiv:2406.03505}.

\bibitem[{Zhang et~al.(2024{\natexlab{c}})Zhang, Wayment-Steele, Brixi, Wang, Kern, and Ovchinnikov}]{zhang2024protein}
Zhidian Zhang, Hannah~K Wayment-Steele, Garyk Brixi, Haobo Wang, Dorothee Kern, and Sergey Ovchinnikov. 2024{\natexlab{c}}.
\newblock Protein language models learn evolutionary statistics of interacting sequence motifs.
\newblock \emph{Proceedings of the National Academy of Sciences}, 121(45):e2406285121.

\bibitem[{Zhao et~al.(2020)Zhao, Liu, Fan, Jiang, Zhao, Yin, and Fu}]{zhao2020simplifying}
Xiaosa Zhao, Kunpeng Liu, Wei Fan, Lu~Jiang, Xiaowei Zhao, Minghao Yin, and Yanjie Fu. 2020.
\newblock Simplifying reinforced feature selection via restructured choice strategy of single agent.
\newblock In \emph{2020 IEEE International conference on data mining (ICDM)}, pages 871--880. IEEE.

\bibitem[{Zou et~al.(2023)Zou, Mo, Li, Cheng, Song, and Xing}]{zou2023linker}
Shuxian Zou, Shentong Mo, Hui Li, Xingyi Cheng, Le~Song, and Eric Xing. 2023.
\newblock Linker-tuning: Optimizing continuous prompts for heterodimeric protein prediction.

\end{thebibliography}

\appendix

\section{Appendix}
\label{sec:appendix}
\subsection{Additional Formulas for DNA-Related Tasks}
\label{app:dnabert}

\subsubsection{Mechanism Explanation (NexLeth): Data Augmentation}
\label{app:nexleth}
In NexLeth, synthetic lethality is explained through rephrased sentences. The method uses:
\begin{equation}
\cos(s, r_i)
= \frac{\mathbf{f}_s \cdot \mathbf{f}_{r_i}}{\|\mathbf{f}_s\|\;\|\mathbf{f}_{r_i}\|},
\tag{A.1}
\end{equation}
to measure the similarity between the original sentence \(s\) and a rephrased one \(r_i\). A maximum log-likelihood objective also appears:
\begin{equation}
P^*
= \arg\max_{P \in \hat{\varepsilon}}
\sum_{j}\log p_{|D|+j},
\tag{A.2}
\end{equation}
where \(\hat{\varepsilon}\) is the set of all possible generated sequences, and \(|D|\) is the length of the input prompt.

\subsection{Additional Formulas for RNA-Related Tasks}
\label{app:pathotme}

\subsubsection{Siamese Network and Domain Adversarial in PathoTME}
Beyond the attention-based MIL in the main text, PathoTME uses a Siamese alignment to integrate genomic embeddings \(z_g\) and WSI embeddings \(p_x\):
\begin{equation}
D_s(p_x, z_g)
= - \frac{p_x \cdot z_g}{\|p_x\|_2\;\|z_g\|_2}.
\tag{A.3}
\end{equation}
A dynamic weight scheduling is introduced to balance losses:
\begin{equation}
\lambda_p
= \frac{2}{1+\exp(-\gamma \, p)} - 1,
\tag{A.4}
\end{equation}
where \(p\) is training progress. The final loss:
\begin{equation}
L_{\text{TOT}}
= (1-\lambda_p) L_S + \lambda_p L_D.
\tag{A.5}
\end{equation}

\subsection{Additional Formulas for Protein-Related Tasks}
\label{app:linker}

\subsubsection{Linker-Tuning: Weighted Distogram Loss}
Below we provide the extended version of weighted distogram losses omitted in Section 2.4.3:
\begin{equation}
\begin{split}
L(x, y, l)
= &L_1(x_A, y_A)
+ L_1(x_B, y_B)\\
&+ \lambda\, L_2(x, y, l).
\end{split}
\tag{A.6}
\end{equation}
The intra-chain loss for chain \(A\):
\begin{equation}
\begin{split}
L_1(x_A, y_A)
= &-\frac{2}{N_A(N_A+1)}\\
&\sum_{i=1}^{N_A}\sum_{j \ge i}^{N_A}\sum_{b=1}^{64} 
y_{ijb}\,\log(p_{ijb}),
\end{split}
\tag{A.7}
\end{equation}
and similarly for chain \(B\). The inter-chain loss:
\begin{equation}
\begin{split}
L_2(x,y,l)
= &-\frac{1}{N_A N_B}\\&
\sum_{i=1}^{N_A}\sum_{j=1}^{N_B}\sum_{b=1}^{64}y_{ijb}\,\log(p_{ijb}).
\end{split}
\tag{A.8}
\end{equation}

\subsubsection{Diffusion Process in InterDiff}
\begin{equation}
q(M_t \mid M_{t-1})
= \mathcal{N}\!\bigl(M_t \,\mid\, \alpha_t M_{t-1}, \sigma_t^2 I\bigr).
\tag{A.9}
\end{equation}
Also, the denoising step can be expressed by:
\begin{equation}
p(x_{t-1} \mid x_t, x_0)
= \mathcal{N}\!\bigl(x_{t-1}\,\mid\,\mu_t(x_t, x_0), \sigma_t^2 I\bigr).
\tag{A.10}
\end{equation}

\subsubsection{Cross-Attention and Optimization in InterDiff}
\begin{equation}
\begin{split}
h_l^{(L)}, x_l^{(L)}
= &\text{CrossAttention}\bigl(cxt,\tilde{d}_{ij},\\
&d_{ij},h_{l-1}^{(L)}, x_{l-1}^{(L)}, x_{l-1}^{(P)}\bigr).
\end{split}
\tag{A.11}
\end{equation}
The final objective is a weighted sum of losses:
\begin{equation}
L
= \lambda_x L_x
+ \lambda_h L_h
+ \lambda_c\,\text{Cls}\bigl(h^{(P)}\bigr).
\tag{A.12}
\end{equation}

\subsubsection{Conformation-Aware Protein Models (ConfProtein)}
A customized attention mask matrix \(M_{ij}\) ensures unidirectional prompt-to-sequence flow:
\begin{equation}
\begin{split}
M_{ij}=\begin{cases}
0, &(1 \leq i \leq m \text{ and } m < j \leq m+n) \\ &\text{ or } (1 \leq i,j \leq m \text{ and } i \neq j) \\
1,& \text{otherwise}.
\end{cases}
\end{split}
\tag{A.13}
\end{equation}
The final hidden state:
\begin{equation}
h
= g\!\Bigl(
\text{softmax}\!\Bigl(\frac{QK^\top}{\sqrt{d}}\cdot M \Bigr)\cdot V
\Bigr).
\tag{A.14}
\end{equation}
The total loss:
\begin{equation}
L
= L_C + \lambda\, L_I,
\tag{A.15}
\end{equation}
where \(L_C\) is knowledge conservation loss, \(L_I\) is injection loss.

\subsubsection{Prompt-Based Protein Reasoning (ProLLM)}
Protein-protein interaction can be notated as \((p_n,p_m,R_i)\). 
Signal transduction can be written as:
\begin{equation}
\text{Rc}\xrightarrow{\text{activate}}\text{Sp}\xrightarrow{\text{inhibit}}\text{Ep}\xrightarrow{\text{response}}M,
\tag{A.16}
\end{equation}
where \(\text{Rc, Sp, Ep, M}\) are distinct proteins or messengers. 
Embedding replacement formula:
\begin{equation}
\text{P}_{\text{seq}}
\xrightarrow{\text{ProtTrans}}
\mathbf{P}_{\text{emb}}
\in \mathbb{R}^{1\times1024}.
\tag{A.17}
\end{equation}

\subsection{Additional Formulas for Drug-Related Tasks}
\label{app:drug}

\subsubsection{HGTDP-DTA: Graph Embedding and Multi-View Fusion}
The original text describes constructing an affinity graph \(\Omega^+\) or \(\Omega^-\). For instance,
\begin{equation}
H^{\text{aff}+}_{d_i}
= \Omega^{+}(G^{+}), 
\quad
H^{\text{aff}-}_{d_i}
= \Omega^{-}(G^{-}),
\tag{A.18}
\end{equation}
\begin{equation}
z^{\text{proj}}_{d_i}
= W_d\!\Bigl[H^{\text{mol}}_{d_i}\;\|\;H^{\text{aff}+}_{d_i}\;\|\;H^{\text{aff}-}_{d_i}\Bigr].
\tag{A.19}
\end{equation}
A dynamic prompt generation then yields \(p_{d_i}\) or \(p_{\text{aff}}\).

\subsubsection{Molecular Design (Latent Prompt Transformer)}
\begin{equation}
p_\beta(x \mid z)
= \prod_{t=1}^T p_\beta\bigl(x^{(t)} \mid x^{(0)}, \dots, x^{(t-1)}, z\bigr),
\tag{A.20}
\end{equation}
\begin{equation}
p_\gamma(y \mid z)
= \frac{1}{\sqrt{2\pi\sigma^2}}
\exp\!\Bigl(-\frac{\bigl(y - s_\gamma(z)\bigr)^2}{2\sigma^2}\Bigr).
\tag{A.21}
\end{equation}
The training objective might combine these two likelihoods and optionally incorporate a property-driven gradient update.

\subsection{Literature Tree}

\begin{figure*}[h!]
    \centering
\rotatebox{-90}{
\tikzset{
    basic/.style  = {draw, text width=5mm, align=center, font=\tiny, rectangle},
    root/.style   = {basic, rounded corners=0pt, thin, minimum height=0.8cm,align=center, fill=cyan!10, font=\tiny},
    tnode/.style = {basic, thin, align=left, fill=green!10, text width=6cm, align=center, font=\tiny},
    xnode/.style = {basic, thin, rounded corners=0pt, align=center, fill=yellow!10, text width=4cm, font=\tiny},
    wnode/.style = {basic, thin, align=left, fill=orange!10,minimum height=0.8cm, text width=12cm, align=center, font=\tiny},
    edge from parent/.style={draw=grey, edge from parent fork right}
}
 
\begin{forest} for tree={
    scale=0.37,
    grow=east,
    growth parent anchor=west,
    parent anchor=east,
    child anchor=west,
    anchor=center,
    s sep=0.5mm,
    edge path={\noexpand\path[\forestoption{edge},->, >={latex}, line width=1.0pt] 
         (!u.parent anchor) -- +(7pt,0pt) |-  (.child anchor) 
         \forestoption{edge label};},
}
[\rotatebox{90}{\textbf{Prompting Applications in Biological Sequences}}, basic, root,anchor=center, l sep=6mm,
    [\textbf{The Significance of AlphaFold and ESM}, xnode, l sep=6mm,
        [ESM Series \cite{hayes2025simulating}, tnode, l sep=6mm,
            [{Output: Protein Representation / Structure Prediction / Functional Annotation / Sequence Generation / Mutagenesis, etc.  },wnode]
            [{Prompt: Mask language modeling + Specific biological tasks.},wnode]
            [{Input: Amino acid sequence of proteins},wnode]]
        [AlphaFold \cite{abramson2024accurate}, tnode, l sep=6mm,
            [{Output: Three-dimensional structure of proteins and evaluation metrics. },wnode]
            [{Prompt: Attention Mechanisms + Geometric Constraints.},wnode]
            [{Input: Amino acid sequence of proteins + MSA },wnode]]] 
    [\textbf{Drug Discovery}, xnode,  l sep=6mm,
        [In-Context Learning for Drug Synergy Prediction \cite{edwards2023synergpt}, tnode,  l sep=6mm,
            [{Output: Predictive value of drug synergistic effect / Inverse designed molecular structur. },wnode]
            [{Prompt: Known drug synergistic relationships, dynamically selected by Random, Graph, Unknown-First.},wnode]
            [{Input: Drug pairs and cell lines},wnode]]
        [Latent Prompt Transformer \cite{kong2024dual}, tnode,  l sep=6mm,
            [{Output: Molecule string + Target attribute prediction. },wnode]
            [{Prompt: Potential vectors as dynamic prompts + Gradual Distribution Shifting.},wnode]
            [{Input: Molecular strings + Chemical/ Biological properties to be optimized.},wnode]]
        [HGTDP-DTA \cite{xiao2024hgtdp}, tnode,  l sep=6mm,
            [{Output:Predictive value for drug-target binding affinity. },wnode]
            [{Dynamic prompt generation: Generate context-specific hints for each drug-target pair.},wnode]
            [{Input: Molecular map of the drug molecule and sequence or residue map of the target protein.},wnode]
        ] ]
    [\textbf{Protein Sequences}, xnode,  l sep=6mm,
        [ProLLM \cite{jin2024prollmproteinchainofthoughtsenhanced}, tnode, l sep=6mm,
            [{Output:Types of protein interactions. },wnode]
            [{Prompt: Protein Chain of Thought, translating signaling pathways into natural language prompts. },wnode]
            [{Input: Protein chain relationships, sequence data and signaling pathway structures},wnode]]
        [ConfProtein \cite{zhang2022promptguidedinjectionconformationpretrained}, tnode, l sep=6mm,
            [{Output:Predictive outcomes (categorical labeling/regression capability) and related assessments},wnode]
            [{Sequence prompts and interactive conformational prompts: capture sequence information and conformational information. },wnode]
            [{Input: Amino acid sequence of proteins / Task: [Determine if two proteins are interacting].},wnode]]
        [PromptMSP \cite{gao2024proteinmultimerstructureprediction}, tnode, l sep=6mm,
            [{Output:Link probability and 3D structure.},wnode]
            [{Prompt: Query chain pairs are combined with the generated virtual nodes to form a 4-node graph.},wnode]
            [{Input: Embedding vectors of protein chains as + assembled multimer graphs + precomputed dimer structures.},wnode]]
        [InstructProtein \cite{wang2023instructprotein}, tnode, l sep=6mm,
            [{Output:Functional Description/Protein Sequences for Specific Functions.},wnode]
            [{Prompt: Task instructions built from the knowledge graph.},wnode]
            [{Input: Protein Sequence / "Design a protein for [some function]".},wnode]]
        [Linker-Tuning \cite{zou2023linker}, tnode, l sep=6mm,
            [{Output:Three-dimensional structure of protein complexes.},wnode]
            [{Prompt: Learnable continuous Linker },wnode]
            [{Input: Amino acid sequence of proteins.},wnode]]
        [InterDiff \cite{wu2024guided}, tnode, l sep=6mm, 
            [{Output:Specific molecular structures or molecular fragments.},wnode]
            [{Interaction Prompt: Types of interactions: $\pi$-$\pi$ interactions, cation-$\pi$ interactions, hydrogen bonding interactions, halogen bonding interactions},wnode]
            [{Input: Coordinates and types of protein and ligand atoms.},wnode]
        ] ]
    [\textbf{RNA Sequences}, xnode,  l sep=6mm,
        [GPTCellType \cite{hou2024assessing}, tnode, l sep=6mm, 
            [{Output: Cell type annotation results.},wnode]
            [{Prompt: Identify cell types of [TissueName] cells by a,b,c},wnode]
            [{Input: List of genes [a,b,c] + name of tissue [TissueName].},wnode]]
        [PathoTME \cite{meng2024genomicsguided}, tnode,l sep=6mm,
            [{Output: Classification results for TME subtypes.},wnode]
            [{Dynamic WSI prompt: Add an additional set of learnable “cue vectors” to the image feature vectors.},wnode]
            [{Input: Whole slice images.},wnode]
        ] ] 
    [\textbf{DNA Sequences}, xnode, l sep=6mm,
        [PLPMpro \cite{li2023plpmpro}, tnode, l sep=6mm,
            [{Output: Classification of DNA sequences as promoters or not}, wnode]
            [{Prompt: k-mer treatment combined with soft templates: [soft][soft][MASK] ATCGAA TCGAGA CGAGA...},wnode]
            [{Input: DNA sequence: ATCGAAAGA...},wnode]]
        [NexLeth \cite{zhang2024prompt}, tnode,l sep=6mm,
            [{Output: A natural language explanation of the mechanism of synthetic lethality of gene pair (a,b)},wnode]
            [{Command prompt: “Explain the mechanism of synthetic lethality of a and b” \texttt{+} KG prompt: “a and b may share the \(\_\_\_\)function”},wnode]
            [{Input: Gene pair (a,b)},wnode]
         ]] ]
\end{forest}
}
    \caption{Literature taxonomy of LLMs in computational biology.}
    \label{fig:lit_surv}
\end{figure*}
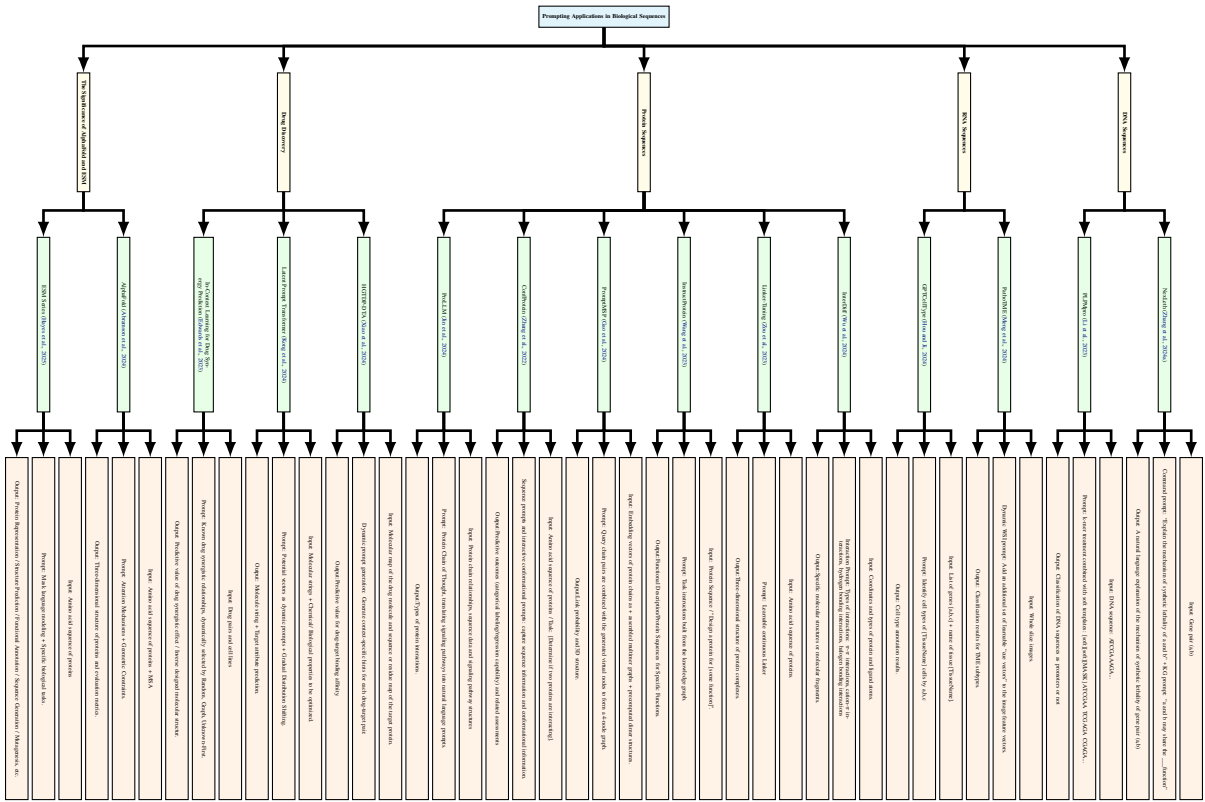

\end{document}